\documentclass{article}
\usepackage{paper}
\usepackage{float}
\usepackage{times}
\usepackage[utf8]{inputenc}
\usepackage[small]{caption}
\usepackage{graphicx}
\usepackage{amsmath}
\usepackage{amsthm}
\usepackage{booktabs}
\usepackage{url}
\usepackage[caption=false]{subfig}
\usepackage{xr}
\usepackage{stfloats}
\newcommand{\tabincell}[2]{\begin{tabular}{@{}#1@{}}#2\end{tabular}}

\title{Motif Difference Field: A Simple and Effective Image Representation of Time Series for Classification
}

\author{
Yadong Zhang \textnormal{and} 
Xin Chen\\
\affiliations
Center of Nanomaterials for Renewable Energy, \\ State Key Laboratory of Electrical Insulation and Power Equipment, \\School of Electrical Engineering, \\ Xi'an Jiaotong University, Xi'an 710054, Shaanxi, China 
\emails \{zhangyadong@stu.xjtu.edu.cn \and 
xin.chen.nj@xjtu.edu.cn
\}
}

\begin{document}

\maketitle

\begin{abstract}

  Time series motifs play an important role in the time series analysis. The motif-based time series clustering is used for the discovery of higher-order patterns or structures in time series data. Inspired by the convolutional neural network (CNN) classifier based on the image representations of time series, motif difference field (MDF) is proposed. Compared to other image representations of time series, MDF is simple and easy to construct. With the Fully Convolution Network (FCN) as the classifier, MDF demonstrates the superior performance on the UCR time series dataset in benchmark with other time series classification methods. It is interesting to find that the triadic time series motifs give the best result in the test. Due to the motif clustering reflected in MDF, the significant motifs are detected with the help of the Gradient-weighted Class Activation Mapping (Grad-CAM). The areas in MDF with high weight in Grad-CAM have a high contribution from the significant motifs with the desired ordinal patterns associated with the signature patterns in time series. However, the signature patterns cannot be identified with the neural network classifiers directly based on the time series.

\end{abstract}

\section{Introduction}

Time series forecasting and classification are important techniques in the understanding of the varieties of dynamics in Science and Engineering. The time series data exist extensively in human daily life and industrial activities, such as health care, transportation, energy, security, finance, {\it etc}. Time series analysis including prediction, anomaly detection, and classification and {\it etc} attracts considerable attention. The classification of time series is an important and challenging problem in data mining. The learning representations and time series classification become more and more important with the development of the Internet of Things (IoT) \cite{IoT_2014} and 5G. Many algorithms have been developed for the problem of time series classification. 


Recently, deep learning has seen a lot of successful cases and applications in the field such as physics, chemistry,  natural language processing, complex network, computer vision, and {\it etc.}. Convolutional neural network (CNN) \cite{CNN_1998} has seen a lot of successes in computer vision such as image classification. Beyond CNN, the recent development in neural networks such as Fully Convolution Network (FCN) \cite{FCN_2017}, ResNet \cite{ResNet_2016}, ScatNet \cite{ScatNet_2013}, AlexNet \cite{AlexNet_2012}, and {\it etc} has been applied in the image classification and segmentation. Motivated by the development, particularly the recent applications of deep learning neural networks for the time series classification \cite{TS_FCN_2016} and image encoding of time series \cite{GAF_MTF_2015}\cite{RP_1987}\cite{EEG}. Based on the raw time series, FCN \cite{TS_FCN_2016} achieves very good performance as classification. The image encoding method Gramian Angular Summation/Difference Fields (GASF/GADF) and Markov Transition Fields (MTF) \cite{GAF_MTF_2015} are proposed. The Tiled CNN \cite{TiledCNN} using GASF/GADF can classify the time series very well while FCN using the raw time series still performs better. The Recurrence Plots \cite{RP_1987} are used for the reconstruction of phase spaces of time series. Also, CNN using the Recurrence Plots \cite{Recurrence_plot} is proposed for the classification of time series. The classifier of CNN and the stacked auto-encoder (SAE) using the time-frequency-domain images \cite{EEG} are also proposed to classify EEG Motor Imagery signals. 

Although the methods such as Recurrence Plots and GASF/GADF do generate distinct image patterns, it remains unclear how to locate the significant temporal motif patterns with the encoding images of time series. Including the high-order temporal structural information in the image encoding will improve the classification accuracy when the time series has the signature temporal patterns.

Time series motifs are the temporal structures in time series. Temporal shapes and motif occurrence probabilities provide the information for the high-order structure. Permutation entropy \cite{PE_2002} based on the time series motif ordinal patterns  has been successfully used for time series complexity measurement, dynamic system characterization, stock market analysis, {\it{etc}}. The triadic time series motif is used for the classification \cite{triadic_classification_2019}. In addition, Symbolic Fourier Approximation (SFA) and Bag-of-Words are the essential components to represent the time series motifs in the Bag-of-SFA-Symbols (BOSS) model \cite{BOSS_2015}. 

Having an image encoding scheme to generate a good time series representation can help us learn the structures in the time series. We propose an imaging encoding scheme called Motif Difference Field (MDF) for the time series based on the motifs of different lengths, which is simple and easy to implement.
We apply FCN to classify the MDF images on 20 datasets of UCR Archive \cite{UCR_2015} and the triadic time series motifs achieve the best performance compared with other image-based neural network classifiers. Different from the known detection approaches \cite{motif_2006} for the significant motifs,, we identify the motif patterns of time series using the Gradient-weighted Class Activation Mapping (Grad-CAM) \cite{Grad_CAM_2017} based on the MDF images and FCN. This can lead to finding the signature patterns in the time series for the classification. A case in the TwoPatterns dataset is used to demonstrate how the Grad-CAM can extract the significant motif patterns from the time series. Furthermore, it can make us comprehend why the MDF image works for time series classification.







\section{Motif Difference Field} 


We introduce the simple motif-based framework for encoding time series as images. 
Given a time series $X=(x_t, t=1,2,3,\cdots,T)$, the sequence of the length $n$ time series motifs is defined as 
\begin{equation}
  \mathbf{M}^n=\{M^n_s, s=1,2,3,\cdots, T-n+1\}
\end{equation}
where $M^n_s$ is the length $n$ time series motif, $1 < n\leq T$, $s$ is the initial time index of the motif and $M^n_s = (x_{t},t=s,s+1,s+2,\cdots,s+(n-1))$. 
Typically, $M^2_s=(x_s, x_{s+1})$ is the dual motif, $M^3_s=(x_s, x_{s+1}, x_{s+2})$  the triadic motif (triad), and $M^4_s=(x_s, x_{s+1}, x_{s+2}, x_{s+3})$ the quad motif. 

In the current definition of the motifs, the displacement of the neighboring points in motifs is unit 1. The large displacements are also important in detecting the temporal structural patterns of time series in the long-range.
By generalizing the motifs with arbitrary displacements, the sequence of generalized motifs is defined as
\begin{equation}
  \mathbf{M}_d^n = \{ M^n_{d,s}, s=1,2,3,\cdots, T-(n-1)d \}
\end{equation}
where $M^n_{d,s}= (x_{t}, t=s,s+d,s+2d,\cdots, s+(n-1)d)$, $d$ is the integer displacement and $ 1 \leq d \leq  d_{max}$ where $d_{max} = \lfloor (T-1)/(n-1) \rfloor$. 

The ordinal patterns of motifs are used in identifying the complexity of time series based on comparing the neighboring values \cite{PE_2002}. In addition to the ordinal patterns, the relative amplitude in motifs have more structural information. Hence, the collection of the sequences of motif differences are defined as,
\begin{equation}
\label{equ:dM}
\mathbf{dM}^n_d=\{dM^n_{d,s}, s=1,2,3,\cdots, T-(n-1)d\}
\end{equation}
where $dM^n_{d,s}=\{x_{s+d}-x_s, x_{s+2d}-x_{s+d}, \cdots, x_{s+(n-1)d}-x_{s+(n-2)d}\}$. 
Since the lengths of $\mathbf{dM}^n_d$ are different, a new sequence $\mathbf{I}^n_d$ is constructed. By assigning $\mathbf{I}^n_d$ to be $\mathbf{dM}^n_d$ initially, $\mathbf{I}^n_d$ is appended with $\mathbf{0}$ recursively until $T-(n-1)$ , where length of $\mathbf{0}=\{0,0,0\cdots,0\}$ is $n-1$. In other words, $\mathbf{I}^n_d = \{I^n_{d,s}, s=1,2,3,\cdots, T-(n-1)\}$ is denoted where
\begin{equation}
  I^n_{d,s} = 
  \begin{cases}
    dM^n_{d,s},  &\; if\; 1 \leq s \leq T-(n-1)d \\
    \mathbf{0} , &\; if\; T-(n-1)d <s \leq T-(n-1)
  \end{cases}
\end{equation}

Then, the motif difference field (MDF) is defined as the collection of $\mathbf{I}^n_d$  with different displacement $d$, 
\begin{equation}
  \mathbf{MDF}^n=\{\mathbf{I}^n_1, \mathbf{I}^n_2, \cdots, \mathbf{I}^n_{d_{max}}\}
\end{equation}

\begin{figure}[H]
\centering
\includegraphics[width=0.5\textwidth]{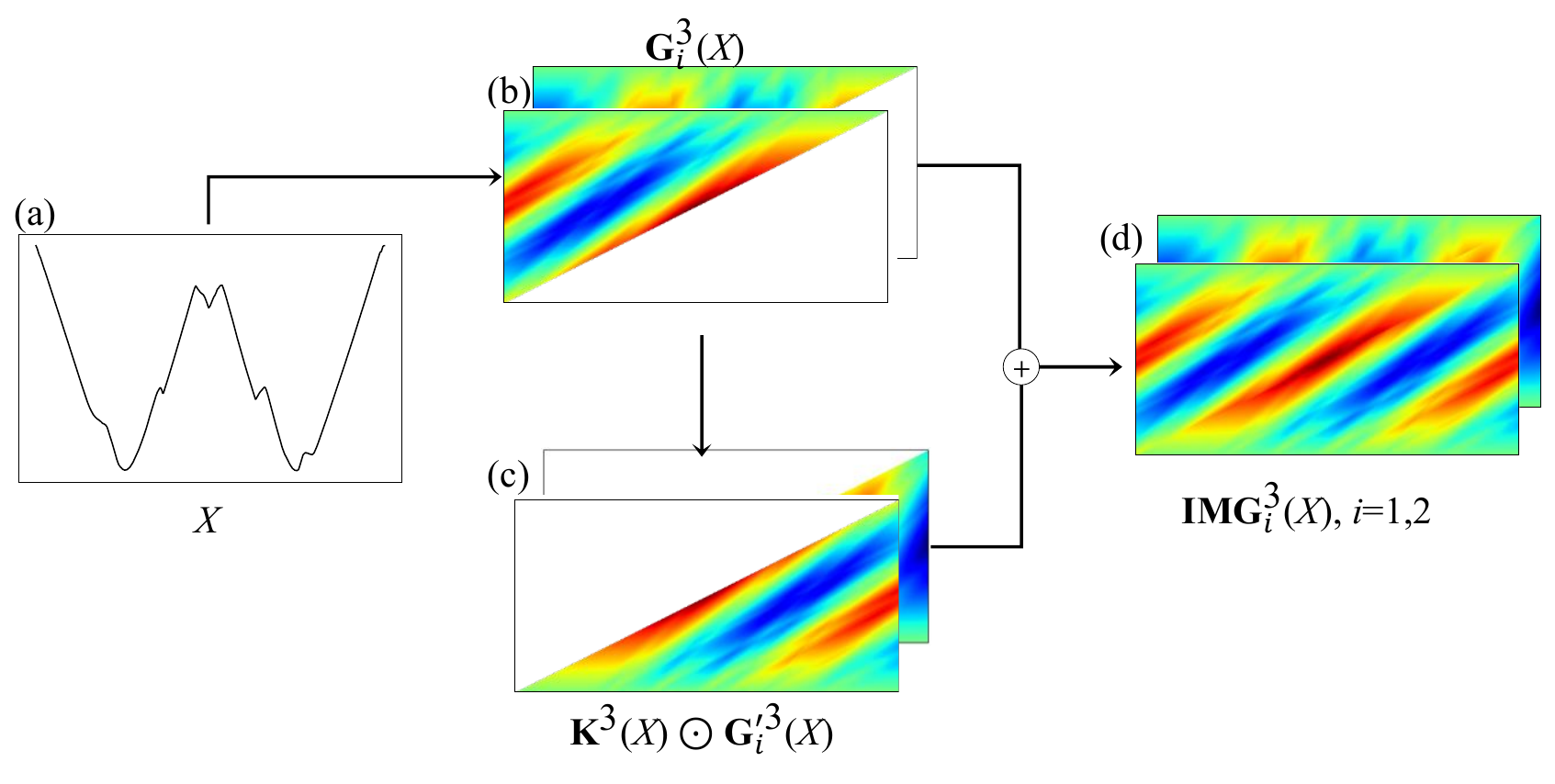}
\caption{The triadic MDF images of a time series instance in the Fish dataset. The blank spaces in the (b) and (c) indicate the $\mathbf{0}$ regions. }
\label{fig:MDF}
\end{figure}

For the motifs of length $n$, the corresponding MDF has $n-1$ elements in $I^n_{s,d}$. Therefore, we can generate $n-1$ channel images accordingly. For $i^{th}$ channel, the array of the image is defined as,
\begin{equation}
  \mathbf{G}^n_i= [\vec{I}^n_{1}(i), \vec{I}^n_{2}(i), \cdots,\vec{I}^n_{d}(i),\cdots,\vec{I}^n_{d_{max}}(i)]^\top
\end{equation}
where $\vec{I}^n_{d}(i) = [I^n_{d,1}(i),I^n_{d,2}(i),\cdots,I^n_{d,T-n+1}(i)]^\top$ and $1\leq i\leq n-1$. In order to fix the plenty of zeros in $\mathbf{G}^n_i$, each channel of MDF image is defined as,
\begin{equation}
  \mathbf{IMG}^n_i = \mathbf{G}^n_i+ \mathbf{K}^n \odot \mathbf{G'}^n_i 
\end{equation}
where $\odot$ is Hadamard product, $\mathbf{G'}^n_i$ is the 180 degree rotation of $\mathbf{G}^n_i$, $\mathbf{K}^n$ is defined as a masker to prevent the overlap of two array, 
\begin{equation}
  K^n_{d,s} = 
  \begin{cases}
    0,  &\; if\; 1 \leq s \leq T-(n-1)d \\
    1 , &\; if\; T-(n-1)d <s \leq T-(n-1)
  \end{cases}
\end{equation}
Figure~\ref{fig:MDF} shows the procedure of encoding the time series into two triadic MDF images. 

\section{Classify Time Series Using MDF with FCN}

We apply the Fully Convolution Network (FCN) to classify the time series using MDF images, named as MDF-FCN. Our classification method is evaluated on 20 datasets from \cite{UCR_2015}. On pre-split testing datasets, we compare the error rate of MDF-FCN with 3 popular classification methods and 3 competing neural network classification methods: 1NN classifier based on DTW with best wrapping window, Bag-of-SFA-Symbols (BOSS) \cite{BOSS_2015}\cite{bakeoff}, Shapelet Transformation (ST) \cite{ST_2012}\cite{bakeoff}, time series with FCN (TS-FCN) \cite{TS_FCN_2016}, GAF-MTF with Tiled CNN \cite{GAF_MTF_2015} and Recurrence Plots with CNN \cite{Recurrence_plot}.

\begin{table*}[!htbp]
  \centering
  \begin{tabular}{lcc|cccccc|ccc}
  \toprule
  Dataset &  Training &  Test &  DTW$^*$ &   BOSS$^*$ &     ST$^*$ &  TS$^*$ &  GM$^+$ &  RP$^+$ & \tabincell{c}{MDF$^+$\\($n=2$)} &    \tabincell{c}{MDF$^+$\\($n=3$)}  & \tabincell{c}{MDF$^+$\\($n=4$)} \\ 
  \hline
  FiftyWords       &       450 &      455 &     0.242  &  0.295 &  0.295 &   0.321 &    0.301 &           0.260 &  0.235 & 0.178 &  \textbf{0.174} \\
  Adiac            &       390 &      391 &     0.391 &  0.235 &  0.217 &   \textbf{0.143} &    0.373 &0.280  &  0.228 &  0.240 &  0.199 \\
  Beef             &        30 &       30 &     0.467 &  0.200 &  0.100 &   0.250 &    0.233 & \textbf{0.080} &  0.133 &  0.133 &  0.167 \\
  CBF              &        30 &      900 &     0.004 &  0.002 &  0.026 &   \textbf{0.000} &    0.009 & 0.005 &  0.114 &  0.002 &  \textbf{0.000} \\
  Coffee           &        28 &       28 &     0.179 &  \textbf{0.000} &  0.036 &   \textbf{0.000} & \textbf{0.000} & \textbf{0.000} & \textbf{0.000} & \textbf{0.000} &  \textbf{0.000} \\
  ECG200           &       100 &      100 &     0.120 &  0.130 &  0.170 &   0.100 &    0.090 &           \textbf{0.000} & 0.120 &  0.050 &  0.110 \\
  FaceAll          &       560 &     1690 &     0.192 &  0.218 &  0.221 &   \textbf{0.071} &    0.237 &           0.190 & 0.091 &  0.102 &  0.193 \\
  FaceFour         &        24 &       88 &     0.114 &  \textbf{0.000} &  0.148 &   0.068 &    0.068 &           0.000 & 0.114 &  0.091 &  0.114 \\
  Fish             &       175 &      175 &     0.160 &  \textbf{0.011} &  \textbf{0.011} &   0.029 &    0.114 &  0.085 & 0.120 & \textbf{0.011} &  0.029 \\
  GunPoint         &        50 &      150 &     0.087 &  \textbf{0.000} &  \textbf{0.000} &   \textbf{0.000} &    0.080 &  \textbf{0.000} & 0.026 &  0.020 &  0.020 \\
  Lightning2       &        60 &       61 &     0.131 &  0.164 &  0.262 &   0.197 &    0.114 &           \textbf{0.000} &  0.230 & 0.180 &  0.213 \\
  Lightning7       &        70 &       73 &     0.288 &  0.315 &  0.274 &   \textbf{0.137} &    0.260 &           0.260 &  0.397 & 0.247 &  0.301 \\
  OliveOil         &        30 &       30 &     0.167 &  0.133 &  0.100 &   0.167 &    0.200 &           0.110 &  \textbf{0.067} & 0.200 &  0.200 \\
  OSULeaf          &       200 &      242 &     0.384 &  0.045 &  0.033 &   \textbf{0.012} &    0.358 &           0.290 & 0.033 & 0.041 &  0.062 \\
  SwedishLeaf      &       500 &      625 &     0.157 &  0.078 &  0.072 &   \textbf{0.034} &    0.065 &           0.060 &  0.046 & 0.051 &  0.043 \\
  SyntheticControl &       300 &      300 &     0.017 &  0.033 &  0.017 &   0.010 &    0.007 &           \textbf{0.000} &  0.017 & 0.017 &  0.010 \\
  Trace            &       100 &      100 &     0.010 &  \textbf{0.000} &  \textbf{0.000} &   \textbf{0.000} &    \textbf{0.000} &  \textbf{0.000} &         \textbf{0.000} &  \textbf{0.000} &  \textbf{0.000} \\
  TwoPatterns      &      1000 &     4000 &     0.002 &  0.007 &  0.045 &   0.103 &    0.091 &           0.170 & 0.003 & \textbf{0.000} &  0.001 \\
  Wafer            &      1000 &     6164 &     0.005 &  0.005 &  \textbf{0.000} &   0.003 &    \textbf{0.000} & \textbf{0.000} & 0.004 & 0.004 &  0.004 \\
  Yoga             &       300 &     3000 &     0.155 &  0.082 &  0.182 &   0.155 &    0.196 &           \textbf{0.000} & 0.183 &  0.175 &  0.158 \\
  \hline
  Average  &    - &    - &    0.164 &   0.098 &  0.110 &  0.090 &   0.140 &   0.090 & 0.108 &  \textbf{0.087} &  0.100 \\
  \bottomrule
  \end{tabular}
  \caption{Error rate of benchmarks and MDF-FCN ($n=\{2,3,4\}$) of 20 datasets in UCR archive. DTW is the 1NN classifier based on DTW with the best wrapping window; TS is the FCN classifier based on time series; GM is the Tiled CNN based on GAF-MTF; RP is the CNN classifier based on Recurrence Plots. The symbols $*$ and $+$ represent the raw-time-series-based methods and image-based methods.}
  \label{tab:result-basic}
\end{table*}

\subsection{Fully Convolution Network}
FCN has shown strong performance and efficiency in image segmentation given the pixel-to-pixel category-wise semantic annotation \cite{FCN_2017} and time series classification \cite{TS_FCN_2016} by extracting the features from the 1-D receptive fields. 
\begin{figure}[H]
  \centering
  \includegraphics[width=0.46\textwidth]{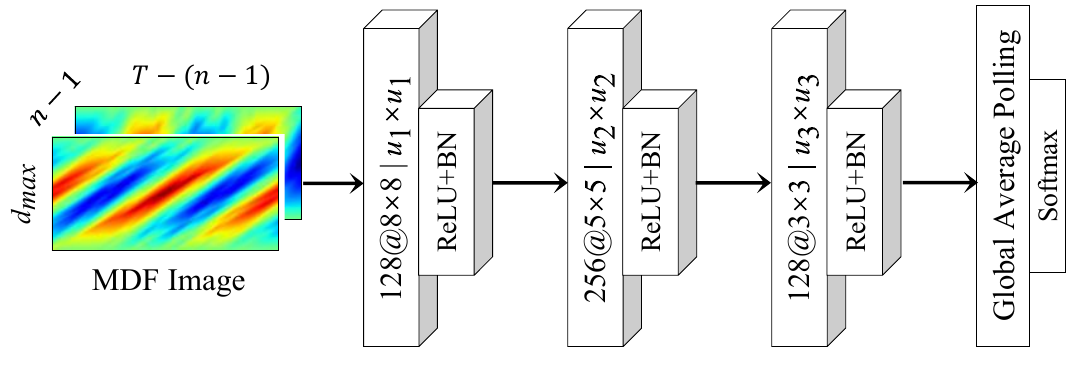}
  \caption{The structure of FCN in the problem setting.}
  \label{fig:network}
\end{figure}
In our problem setting, FCN is structured as the 2-D features extractor of the MDF image. As shown in Figure \ref{fig:network}, the basic block is convolutional layers followed by ReLU activation and batch normalization (BN) layer, 
\begin{equation}
  A_{z+1} = ReLU(BN(\mathbf{W}_z \otimes A_{z} + b_z))
\end{equation}
where $\otimes$ is the convolution operator, $0\leq z\leq3$, $A_0$ is the input MDF image, $A_z$ is the feature map of the $z^{th}$ convolutional layer if $z\neq0$. 
The classification output $y$ comes from the global average pooling layer (GAP) of $A_3$ followed by softmax layer,
\begin{equation}
  h = GAP(A_3)
\end{equation}
\begin{equation}
  y = Softmax(\mathbf{W}_o h + b_o)
\end{equation}
where $y$ is a size $C$ vector, $C$ is the classes number of the dataset. 

The three convolutional layers in FCN have 2-D receptive fields $\{8\times 8, 5\times 5, 3\times 3\}$ and the filter size $\{128, 256, 128\}$. Moreover, the strides of convolution operators affect the efficiency and the overlap size of receptive fields, our experiment conducts the cross-validation to provide
optimal strides $\{u_1\times u_1, u_2\times u_2, u_3 \times u_3\}$. 

\subsection{Experiment Settings}

In our experiment, we use three different MDF images for FCN with the lengths $n\in\{2,3,4\}$. We test the FCN classifier using MDF images on the 20 univariate time series datasets \cite{UCR_2015}. All the datasets have been split into training and testing by default. Each dataset is preprocessed with minmax normalization with the minimum and maximum value of its training datasets. 
Taking categorical cross entropy as loss function, FCN is trained with Adam \cite{Adam_2015} at the learning rate 0.001, $\beta_1=0.9$ and $\beta_2=0.999$. The strides of FCN, $(u_1, u_2, u_3)$, are selected to be $\{(8, 5, 3), (4, 2, 2), (2, 2, 2), (3, 2, 1)\}$ according to 4-fold cross-validation on the training dataset. Finally, we choose the best model that achieves the lowest training loss and report its performance on the test dataset in Table~\ref{tab:result-basic}.

\subsection{Results and Analysis}

We benchmark MDF-FCN with other 6 classifiers. The classification error rates on testing datasets are reported in Table~\ref{tab:result-basic}. The motifs of small lengths, such as $\{2,3,4\}$ in the Table~\ref{tab:result-basic}, can allow to see the long range patterns due to the large possible displacements. Our results show that the triadic MDF performs better than the dual and quad in general. 

Benchmarking with other six classifiers, FCN based on triadic MDF gives the smallest average of error rate. Compared to 3 popular methods, the triadic MDF-FCN performs better on 11 datasets since the high-order patterns of the time series embedded in the MDF images can not be identified by the distance-based and feature-based methods. Compared to the two other image encoding classification methods, triadic MDF-FCN performs better on other 11 datasets and the image encoding in MDF doesn't need complicated parameters and processing. The previous work \cite{TS_FCN_2016} shows that TS-FCN classification method can achieve a nice performance only with the raw time series data. However, TS-FCN can't extract or detect the high-order temporal patterns while MDF-FCN can identify the significant motif ordinal patterns. That's why MDF-FCN outperforms TS-FCN. This will be demonstrated in the next section.

\section{Identify Significant Motif Ordinal Patterns}
\begin{figure*}[th]
  \centering
  \subfloat[A time series instance]{\includegraphics[width=0.45\textwidth]{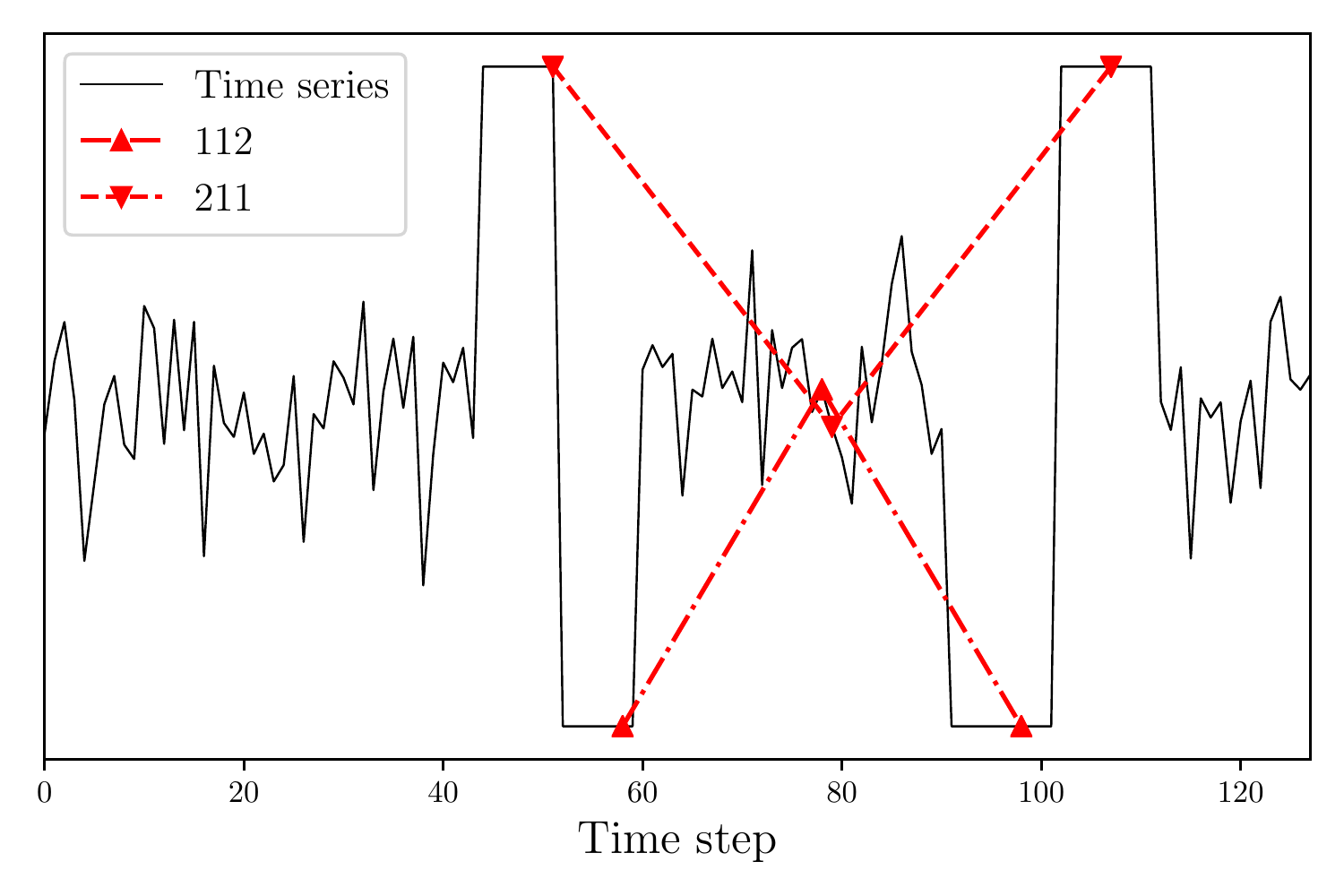}\label{f1}}
  \subfloat[The triadic MDF image]{\includegraphics[width=0.5\textwidth]{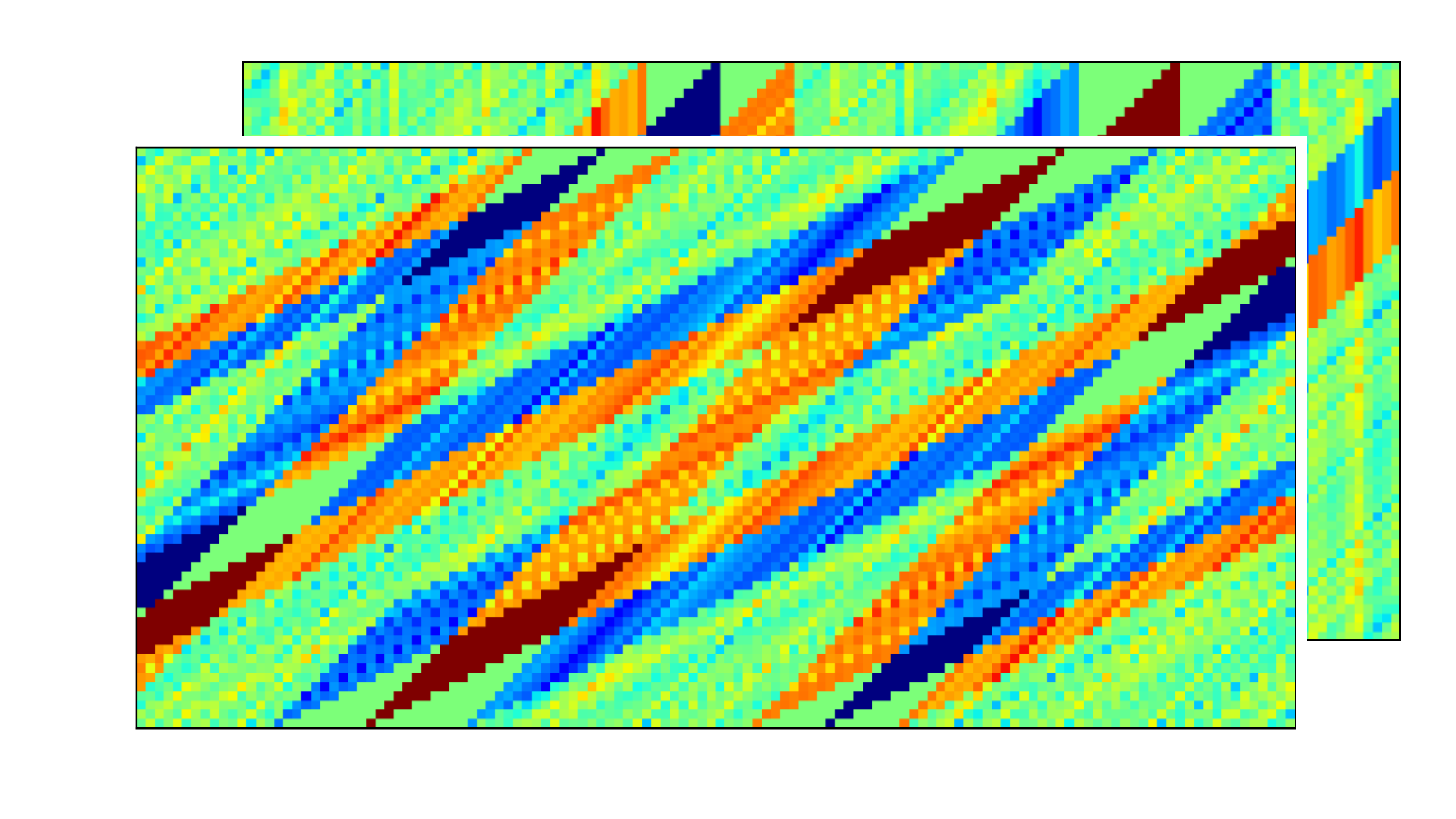}\label{f2}} \\ 
  \subfloat[The Grad-CAM heat-map]{\includegraphics[width=0.5\textwidth]{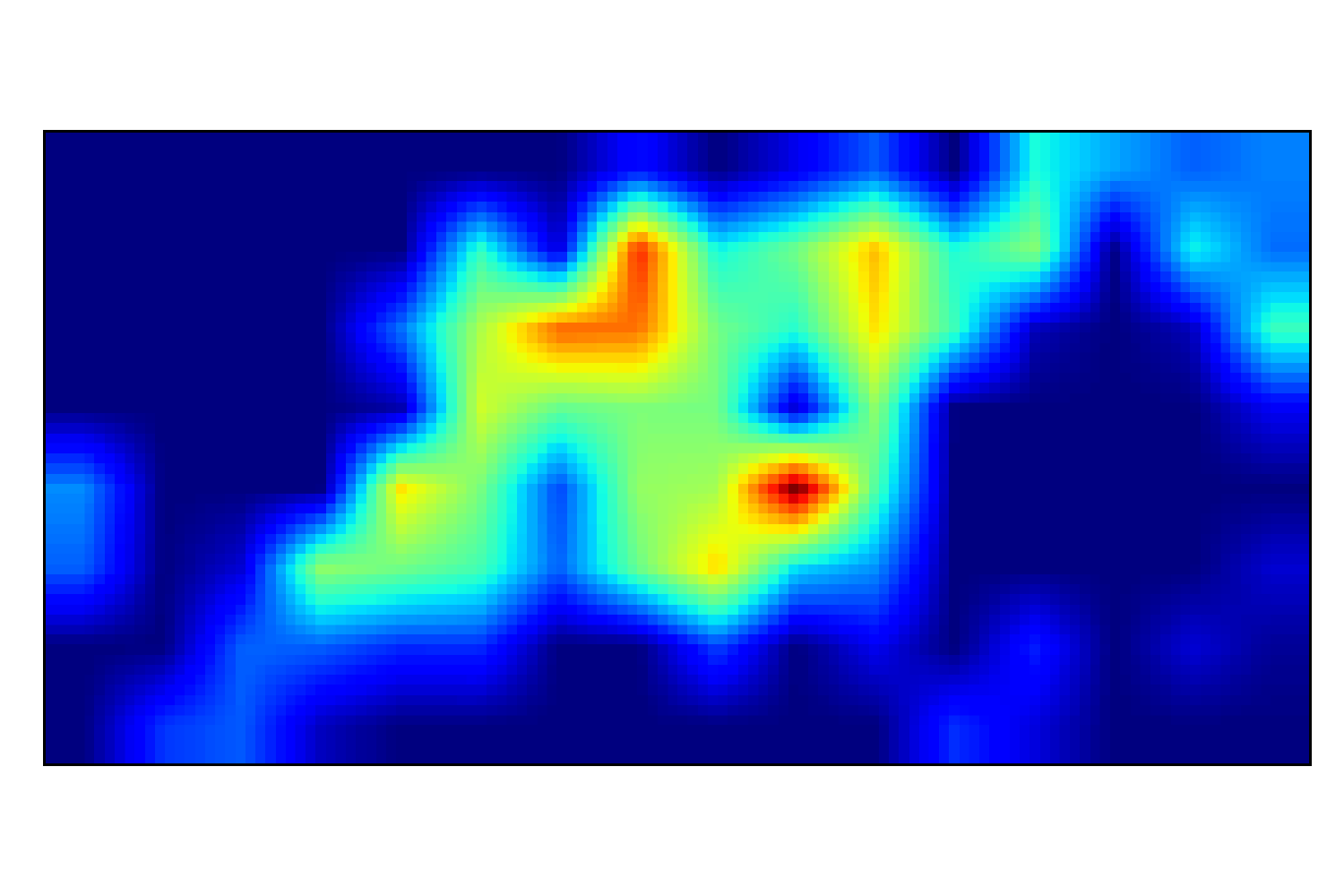}\label{f3}}
  \subfloat[The Symmetrized Grad-CAM heat-map]{\includegraphics[width=0.5\textwidth]{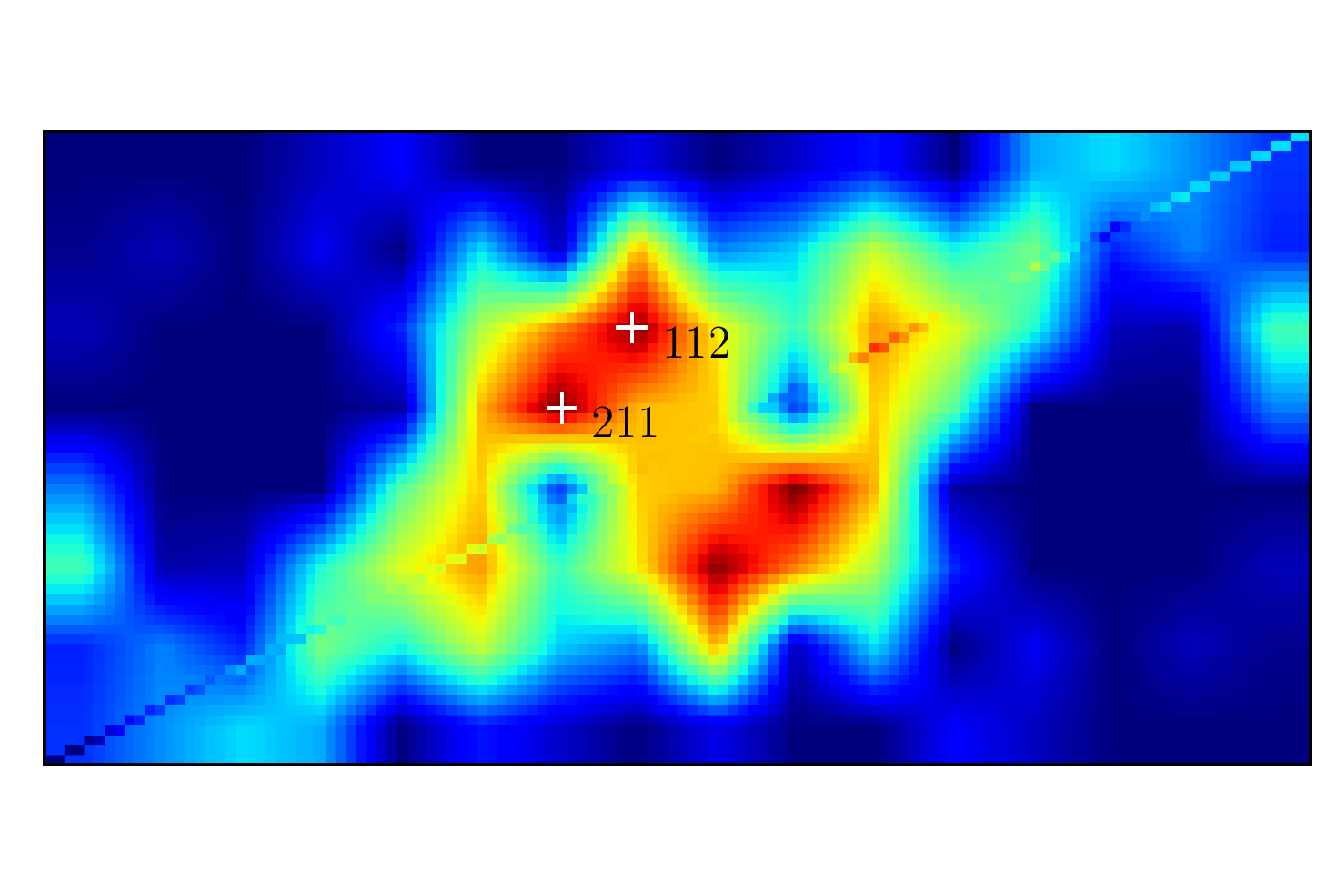}\label{f4}}
  \caption{The Grad-CAM heat-maps of triadic MDF images  of the time series instance in the TwoPatterns dataset. Plot (a) shows the time series instance, Plot (b) the triadic MDF images, Plot (c) the Grad-CAM heat-map and Plot (d) The Symmetrized Grad-CAM heat-map. The two motif patterns of 112 and 211 in Plot (a) are the two peaks in Plot (d) and are the top two significant ordinal patterns according to Equation~\ref{eq:rank}.}\label{fig:GCAM}
\end{figure*} 
The Gradient-weighted Class Activation Mapping (Grad-CAM) \cite{Grad_CAM_2017} is a technique to produce a visual explanation to varieties of CNN model-families.  
In the paper, we will apply Grad-CAM
to detect the significant motifs that are associated with the signature patterns in time series classification.  
The Grad-CAM of FCN using MDF images can be generated by the gradient-based weighted combinations of the feature maps of the last convolutional layer $A_3$ in Figure~\ref{fig:network},
\begin{equation}
  L^c = ReLU(\sum_k(\alpha^c_k A^k_3))
\label{equ:grad-cam}
\end{equation}
where $c$ is the target class of time series, $k$ is channel index of the feature map and $\alpha^c_k=GAP(\partial y^c / \partial A^k_3)$.

\subsection{Symmetrized Grad-CAM}
Given the MDF images of the time series in the target class, we calculate the Grad-CAM, $L^c$, of the last FCN convolutional layer based on Equation \ref{equ:grad-cam}. The image of Grad-CAM will be up-sampled to the size of $d_{max}\times [T-(n-1)]$. For simplicity, the up-sampled Grad-CAM is named as $L^c_{d,s}$. $L^c_{d,s}$ gives the significance of the motif difference $dM^n_{d,s}$ in the MDF images of time series of class $c$. Due to the symmetry in the MDF images as constructed, we enforce the same symmetry by defining the new symmetrized Grad-CAM $L'^c_{d,s}$ with $(L^c_{d,s}+L^c_{d^*,s^*})/2$ where $(d,s)$ and $(d^*,s^*)$ are the two array indices according to the 180 rotation symmetry in $G^n$ and masked $G'^n$. Figure~\ref{f1} shows a time series instance in TwoPatterns testing dataset. Taking the instance as an example, Figure~\ref{f2} shows the two MDF channel images. Figures~\ref{f3} and ~\ref{f4} show the original Grad-CAM and symmetrized Grad-CAM of the MDF image respectively.

\subsection{Significant Motif Ordinal Patterns}
It is helpful to explain the efficiency of MDF image classification with the symmetrized Grad-CAM. Based on the symmetrized Grad-CAM, we can assign significance to the motifs. 
We categorize the motifs of length $n$ according to the ordinal patterns $\{\pi^n_j, j=1,2,3,\cdots,m\}$ \cite{Extend_alphabet}. For dual time series motifs, there are three ordinal patterns; for the triadic time series motifs, there are thirteen ordinal patterns; for the quad time series motifs, there are seventy-three ordinal patterns. Since the triadic MDF-FCN gives the best results in Table~\ref{tab:result-basic}, we will focus on the thirteen ordinal patterns of $\pi^3_j$, 
\begin{equation}
  \begin{split}
    \pi^3_j \in \{111,112,113,122,123,132,211,\\213,221,231,311,312,321\}
  \end{split}
\end{equation}
Accordingly, the collection of the indices of all the $\pi^n_j$ motifs in $\mathbf{MDF}^n$ is defined as,
\begin{equation}
  \begin{split}
    MC_j = \{ (d,s) \vert M^n_{d,s} \in \pi^n_j , 1 \leq s \leq T - (n-1)d,\\ 1 \leq d \leq d_{max}\}
  \end{split}
\end{equation}
The contribution of $\pi^n_j$ motif ordinal patterns can be evaluated according to $E_j$,
\begin{equation}
E_j = \frac{1}{Z_j} \sum_{MC_j} L'^c_{d,s}
\end{equation}
where $Z_j$ is the length of $MC_j$. $E_j$ can be regarded as the significance of $\pi^n_j$ ordinal patterns in the MDF image. 

Based on $E_j$, we can get the ranking of the significance of $\pi^n_j$ patterns,
\begin{equation}\label{eq:rank}
  E_{v(1)} \geq E_{v(2)} \geq \cdots E_{v(j)} \cdots \geq E_{v(m)}
\end{equation}
where $v$ is the mapping of the monotonic ascending index of 1 to $m$ to the index of $\{MC_j, j=1,2,3,\cdots,m\}$. 
We decide which ordinal patterns of the motif are the most important according to the $E_j$ ranking. Given $MC_j$, we can evaluate the contributions of the motifs of $\pi^n_j$ in the symmetrized Grad-CAM. 

\subsection{MDF-FCN {\it v.s.} TS-FCN}
\begin{figure}[H]
\centering
\includegraphics[width=0.45\textwidth]{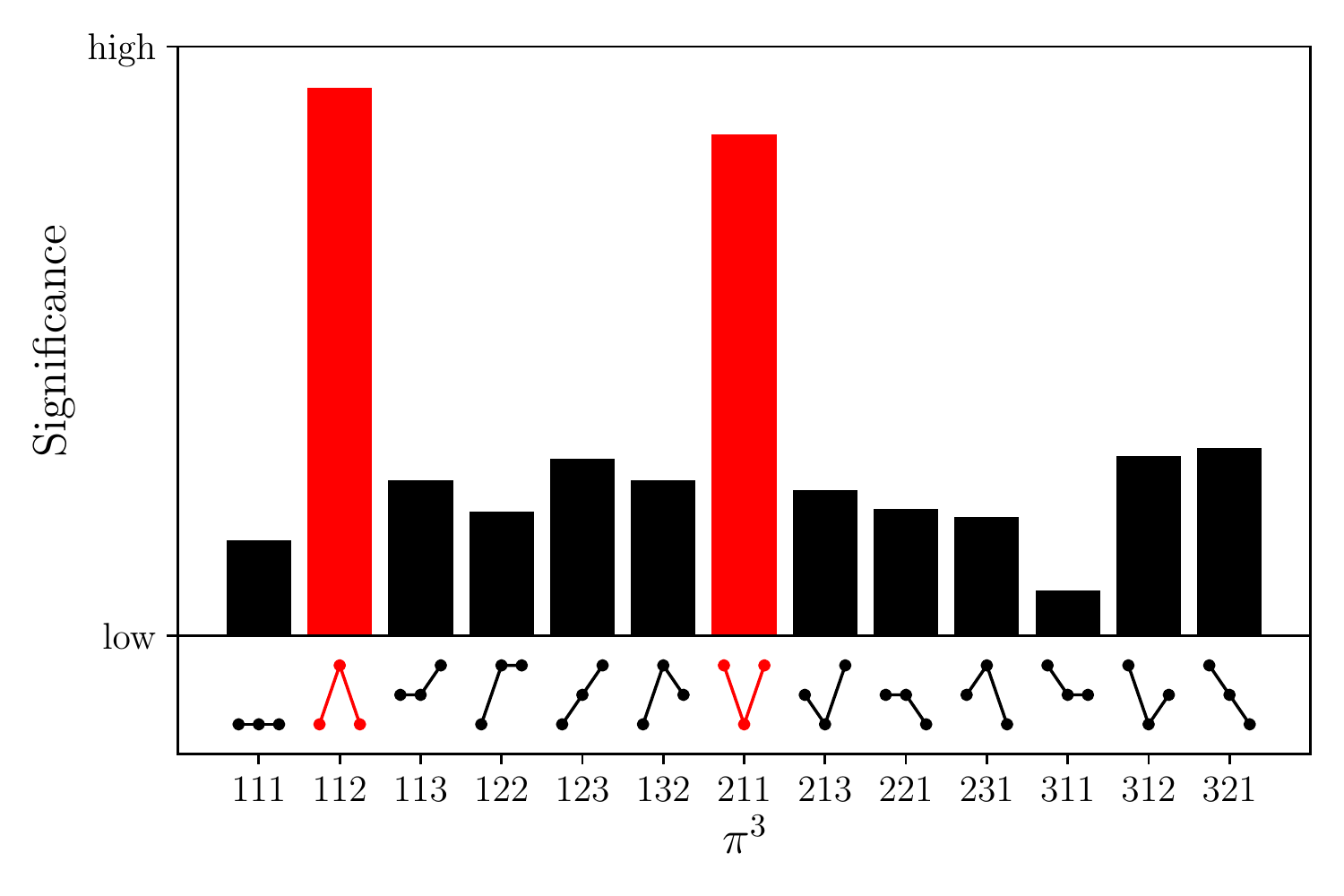}
\caption{Significance of triadic time series motif ordinal patterns $\pi^3_j$. The bottom panel shows thirteen triadic time series ordinal patterns.}
\label{fig:motif-contribution}
\end{figure}

\begin{figure}[H]
  \centering
  \includegraphics[width=0.45\textwidth]{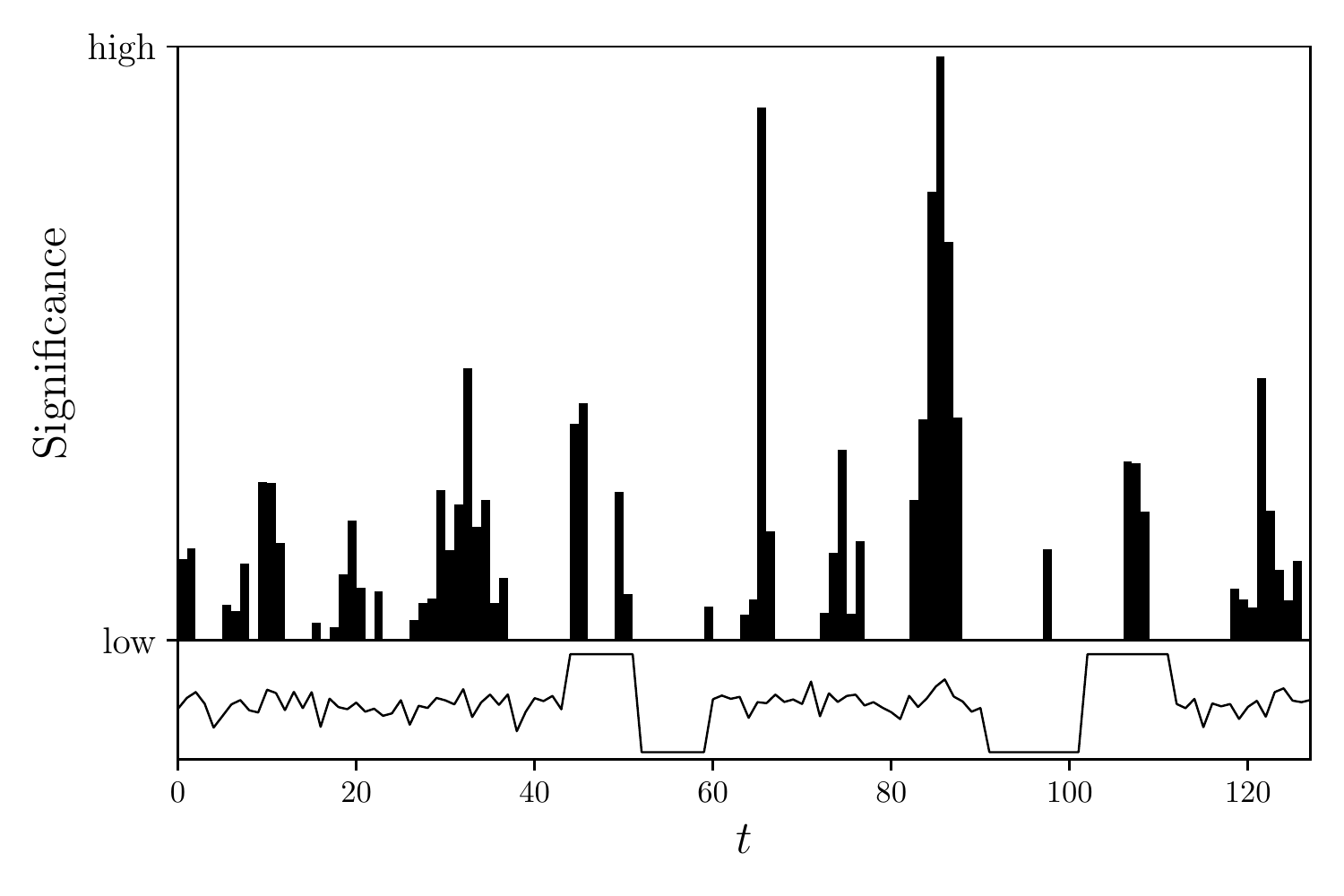}
  \caption{Significance of the data points of the time series calculated by the Grad-CAM of TS-FCN. The bottom panel shows the original time series. }
  \label{fig:TS-freq}
  \end{figure}

In Table~\ref{tab:result-basic}, the triadic MDF images give the best results. Figure \ref{fig:GCAM} show the time series instance of TwoPatterns dataset and its triadic MDF images and Grad-CAM heat-map. In the TwoPatterns dataset, all the time series have the four plateaus which serve as the unique signatures for the classification. We can rank ordinal patterns of triadic motifs according to Equation~\ref{eq:rank}. Figure~\ref{fig:motif-contribution} shows the significance of thirteen $\pi^3_j$ motif ordinal patterns accordingly. It is found that the top two significant motif ordinal patterns are $\{112, 211\}$ corresponding to the order of the four plateaus as shown in Figure~\ref{f1}. Furthermore, the two ordinal patterns correspond to the hottest peaks in the symmetrized Grad-CAM heat-map as shown in Figure \ref{f4}.  At the same time, Grad-CAM for the TS-FCN classifier \cite{TS_FCN_2016} can be calculated, and up-sampled to the size $T$. Since TS-FCN is fed with the raw univariate time series directly, the Grad-CAM in Equation~\ref{equ:grad-cam} is the significance of the individual data points in the time series. Figure~\ref{fig:TS-freq} shows the significance of the data points. The TS-FCN classifier puts much more weight in the fluctuating parts instead of four signature plateaus. This explains why MDF-FCN classifier performs much better than TS-FCN classifier when the time series has the signature of a higher-order structure. 

\section{Conclusion and Future Works}
We propose a framework for encoding time series into MDF images and use FCN as the 2-D feature extractor for classification. Our benchmark study shows that the triadic MDF based on FCN achieves the best results for classification compared to other classification methods. With the help of Grad-CAM, we identify the significant motif ordinal orders that associated with the signature patterns in time series. Our analysis suggests that the MDF images encoding has the following advantages:
\begin{enumerate}
  \item MDF images represent the motif-based temporal structural patterns of time series at the long and short ranges. From this point, it is a more effective classifier than TS-FCN. 
  \item MDF images are much simpler than the Recurrence Plots and the GASF/GADF-MTF.
  \item Grad-CAM based on MDF-FCN can identify the significant ordinal patterns in time series. It helps us comprehend and analyze the time series classification from the motif perspective. 
\end{enumerate}

There is more for us to do to explore how to make the MDF-FCN classifier more efficient in the classification. The future work will include investigating how the lengths of motifs affect the performance of classification. There is more work for us to test the new image classification neural networks in the classification using the image encoding of time series. In addition, how to use the MDF-FCN method for the identification of the higher-order temporal structure in the time series is a promising application. How to classify time series using the motifs of different lengths demands more understanding of the current MDF images. The MDF-FCN framework can be extended for the time series prediction and anomaly detection.
\section*{Acknowledgments}
  Xin Chen acknowledges the financial support from the National Natural Science Foundation of China (Grant No. 21773182 (B030103) ).

\appendix

\clearpage
\bibliographystyle{named}
\bibliography{paper}

\end{document}